\documentclass[conference]{IEEEtran}
\IEEEoverridecommandlockouts

\usepackage{cite}
\usepackage{amsmath,amssymb,amsfonts}
\usepackage{algorithmic}
\usepackage{graphicx}
\usepackage{textcomp}
\usepackage{xcolor}
\usepackage{caption}
\usepackage{subcaption}
\usepackage{hyperref}
\usepackage{balance}
\usepackage{dblfloatfix}
\usepackage{placeins}

\def\BibTeX{{\rm B\kern-.05em{\sc i\kern-.025em b}\kern-.08em
    T\kern-.1667em\lower.7ex\hbox{E}\kern-.125emX}}

\makeatletter

\def\ps@IEEEtitlepagestyle{
  \def\@oddfoot{\mycopyrightnotice}
  \def\@evenfoot{}
}

\def\mycopyrightnotice{
  {\footnotesize\hfill}
  \gdef\mycopyrightnotice{}
}

\newcommand*\titleheader[1]{\gdef\@titleheader{#1}}

\AtBeginDocument{%
  \let\st@red@title\@title
  \def\@title{%
    \bgroup
    \normalfont\large\centering
    \@titleheader\par
    \egroup
    \vskip1.5em
    \st@red@title
  }
}

\makeatother

\makeatletter

\let\old@ps@headings\ps@headings
\let\old@ps@IEEEtitlepagestyle\ps@IEEEtitlepagestyle

\def\confheader#1{%
  \def\ps@headings{%
    \old@ps@headings
    \def\@oddhead{\strut\hfill#1\hfill\strut}%
    \def\@evenhead{\strut\hfill#1\hfill\strut}%
  }%
  \def\ps@IEEEtitlepagestyle{%
    \old@ps@IEEEtitlepagestyle
    \def\@oddhead{\strut\hfill#1\hfill\strut}%
    \def\@evenhead{\strut\hfill#1\hfill\strut}%
  }%
  \ps@headings
}

\makeatother

\begin{document}

\title{CohortHijack: Robustness of Single Cell Annotation to Companion Cell Removal}


\author{
Arash Vashagh$^*$ and Yasmin Vashagh$^\dagger$\\
$^*$ Faculty of Computer Science, University of New Brunswick,
Fredericton, New Brunswick E3B 5A3, Canada\\
$^\dagger$ Farzanegan Amin 2 High School, Isfahan, Iran\\
\{arash.vashagh@unb.ca, yasmin.vashagh@gmail.com\}
}

\maketitle

\begin{abstract}
Many single-cell annotation tools refine an initial cell label using nearby
cells or cluster-level voting. We study whether this refinement can be
manipulated without changing the target cell. We introduce CohortHijack, a
robustness audit that removes selected non-target cells from the query cohort
while preserving the target expression profile, base prediction, and trained
model. We evaluate random and structured removal methods, together with greedy,
multi-start, and beam search, on PBMC3K and Paul15 using logistic regression
and calibrated linear SVM classifiers. Structured removal was consistently
stronger than random removal on Paul15. Multi-start search changed 24.33\% of
linear-SVM targets and 19.67\% of logistic-regression targets while removing a
small fraction of the cohort and keeping mean collateral changes below 0.4\%.
Ablations confirmed that the effect disappeared when neighborhood refinement
was disabled. We also evaluated CellTypist majority voting, where independent
predictions remained unchanged across all evaluations, but refined labels
changed after small companion-cell removals. These findings identify query
cohort composition as a target-preserving attack surface in single-cell
annotation.
\end{abstract}

\begin{IEEEkeywords}
RNA sequencing, cell-type annotation, adversarial robustness,
cohort dependence,
CellTypist
\end{IEEEkeywords}

\section{Introduction}
\label{sec:introduction}

Single-cell RNA sequencing enables the measurement of gene expression at the
level of individual cells and supports the study of complex tissues, immune
populations, developmental processes, and disease states. Modern analysis
pipelines commonly include normalization, dimensionality reduction,
clustering, visualization, and cell-type annotation
\cite{Wolf2018Scanpy,CrossTissueImmuneCell}. Since large datasets may contain
thousands or millions of cells, manual annotation is slow and may vary between
experts. Automated cell-type annotation has therefore become an important part
of single-cell analysis
\cite{Abdelaal2019Comparison,Kiselev2018scmap,Zhang2019CellAssign}.

Existing annotation methods use different forms of prior knowledge and
reference information. Some methods project query cells onto annotated
reference datasets, while others use probabilistic models, marker genes,
integration methods, or consensus predictions
\cite{Kiselev2018scmap,Xu2021Probabilistic,ANDREATTA20213796,
Ergen2024popV}. Reference mapping methods such as Symphony place query cells
within a stable reference representation and transfer known annotations to
new datasets \cite{Kang2021}. Other frameworks combine annotation,
integration, and label refinement to improve predictions across datasets
\cite{10.1093/bib/bbae305,10.1093/bioinformatics/btac140}. These approaches
show that the final annotation of a cell may depend not only on its own gene
expression, but also on the surrounding cells, neighborhood structure, or
cluster-level information.

Cohort information can improve annotation by correcting isolated or uncertain
predictions. Consensus methods combine outputs from several annotation
algorithms, while neighborhood and niche-based methods use nearby cells or
local cellular composition
\cite{Ergen2024popV,Agrawal2024}. However, this dependence also creates a
robustness concern. If the query cohort changes because some cells are removed,
the neighborhood graph, cluster assignments, local class frequencies, and
refined labels may also change. As a result, the same target cell may receive a
different final annotation even though its gene-expression profile remains
unchanged.

Changes in cohort composition can occur during ordinary single-cell analysis.
Low-quality cells may be removed during quality control, rare cells may be lost
during sampling, and large datasets may be reduced before downstream analysis.
Previous work has shown that random subsampling may discard rare populations
or distort the structure of a single-cell dataset. This motivated methods that
select cells more carefully, either to preserve cellular diversity or to retain
cells with high value for downstream analysis
\cite{scSampler,Huang2025scValue}. These studies focus on preserving global
dataset quality. They do not examine whether a small and carefully selected set
of removed cells can change the annotation of a specific unchanged target
cell.

Adversarial attacks provide a useful framework for studying such worst-case
behavior. Early work showed that small input perturbations can cause accurate
machine learning models to output incorrect predictions
\cite{szegedy2013intriguing,goodfellow2014explaining}. Later research has studied attacks against model utility, privacy, and
explainability, together with defenses and target-preserving attacks that
alter external context rather than the target input
\cite{RecentAdvancesinAdversarialAttacks,202607.1022,
Vashagh2026ConformalShift}. In single-cell
analysis, adverSCarial evaluates the vulnerability of RNA-sequencing
classifiers to adversarial perturbations of gene-expression inputs
\cite{10.1093/bioinformatics/btaf168}. Such attacks modify the target cell's
features. In contrast, the vulnerability studied here preserves the target
cell and changes only the surrounding query cohort.

A related idea appears in attacks against graph-based learning. Graph
classifiers may be affected when edges, neighboring nodes, or graph structure
are changed
\cite{pmlr-v80-dai18b,zugner2020adversarial}. These studies demonstrate that a
prediction can depend on relational context rather than only on the target
node's features. However, they do not study single-cell annotation pipelines,
cell-removal constraints, biological neighborhoods, or label-refinement
systems. Hence, the effect of companion-cell removal on an unchanged single-cell
target remains underexplored.

On the other hand, annotation reliability is especially important for cells with uncertain or
ambiguous identities. Recent work has proposed consensus voting, hierarchical
rejection, and training-dynamics analysis to identify low-confidence
annotations, possible labeling errors, and intermediate cell states
\cite{10.1093/bioinformatics/btae128,Karin2024}.

We introduce \emph{CohortHijack}, a robustness audit that removes selected
non-target companion cells and tests whether the target's refined annotation
changes. Unlike feature-based attacks, CohortHijack preserves the target
expression vector, reference labels, classifier, and model parameters.

\begin{figure*}[t]
    \centering
    \includegraphics[width=0.98\textwidth]{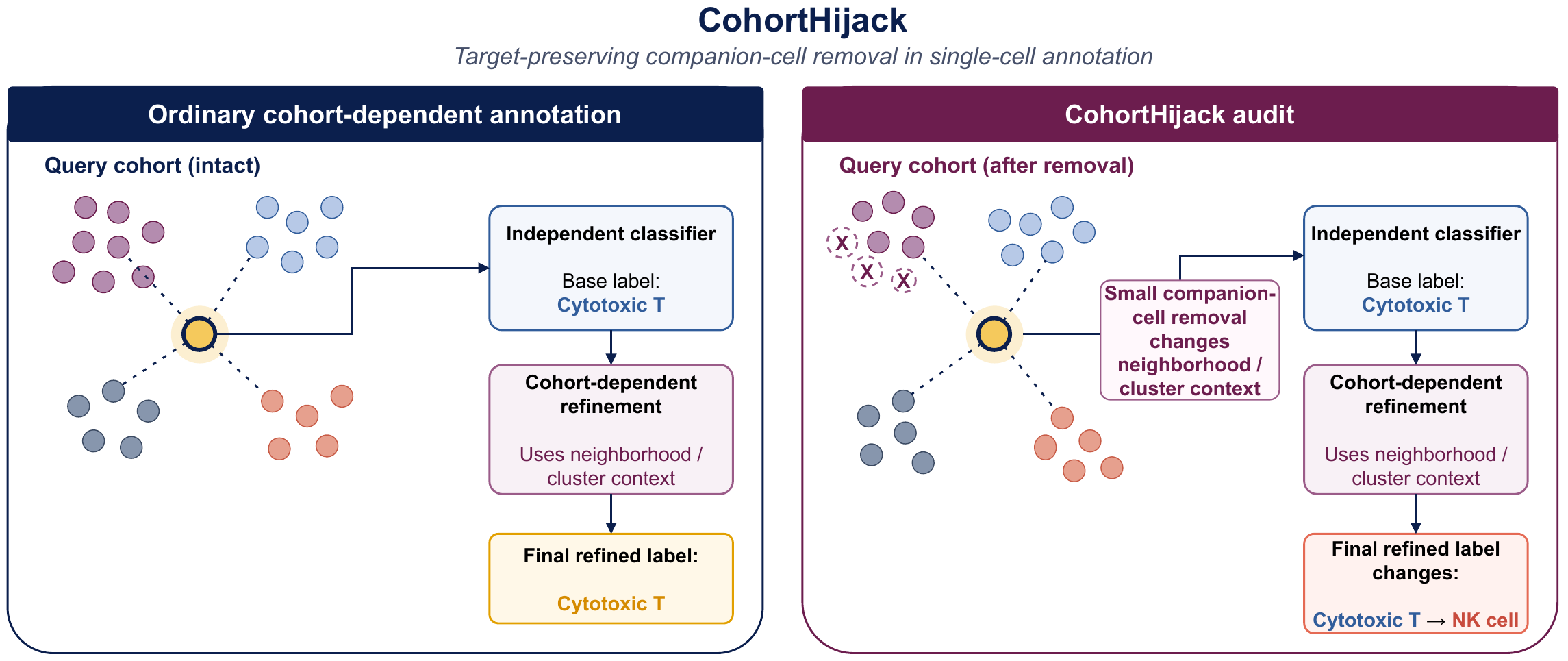}
    \caption{Overview of the CohortHijack threat model. Under ordinary
    annotation, the target cell is classified using its own expression profile
    and contextual information from the complete query cohort. CohortHijack
    removes a small set of non-target companion cells and recomputes the
    cohort-dependent refinement. The target expression vector, base classifier,
    and model parameters remain unchanged, but the final refined annotation may
    change because the neighborhood or clustering context has changed.}
    \label{fig:cohorthijack_overview}
\end{figure*}

We study random removal, nearest-neighbor removal, same-class removal, greedy
search, multi-start greedy search, and beam search. The controlled experiments
use two single-cell datasets, two frozen classifier families, and three random
seeds. We also vary the neighborhood size, context weight, and voting rule to
identify the source of the vulnerability. Finally, we validate the same threat
model using CellTypist majority voting \cite{CrossTissueImmuneCell}, where the target's independent
prediction remains unchanged while its cohort-refined label may change after a
small fraction of companion cells is removed.
Figure~\ref{fig:cohorthijack_overview} illustrates the difference between
ordinary cohort-dependent annotation and the CohortHijack audit.

Our work makes three contributions. First, we define companion-cell removal
as a target-preserving threat to refined single-cell annotation. Second, we
develop structured and search-based removal methods and measure both target
and collateral changes. Third, we validate the threat across controlled
pipelines and CellTypist majority voting.\footnote{Code and reproduction
scripts: \href{https://github.com/arashVsh/CohortHijack}{https://github.com/arashVsh/CohortHijack}}

\section{Method}
\label{sec:method}

This section presents the CohortHijack evaluation framework, including the
annotation pipeline, removal strategies, search methods, evaluation measures,
and CellTypist validation. We use \emph{audit} for the overall evaluation
protocol and \emph{attack} for a removal set selected to change a target label.

\subsection{Problem Setup}
\label{subsec:problem_setup}

Let $\mathcal{D}=\{(\mathbf{x}_i,y_i)\}_{i=1}^{n}$ denote a single-cell
dataset containing $n$ cells. The vector $\mathbf{x}_i\in\mathbb{R}^{d}$
contains the processed features of cell $i$, and $y_i\in\mathcal{Y}$ is its
reference cell-type label. The set $\mathcal{Y}$ contains all cell types
retained after preprocessing. We divide $\mathcal{D}$ into disjoint training
and test sets using a stratified split. The test set is treated as the query
cohort during the robustness audit.

For a selected target cell indexed by $t$, CohortHijack removes a subset
$\mathcal{S}$ of non-target test cells. The target is never included in the
removal set, so $t\notin\mathcal{S}$. The attacked query cohort is
$\mathcal{D}_{\mathcal{S}}=\mathcal{D}_{\mathrm{test}}\setminus\mathcal{S}$.
The target feature vector, trained classifier, and base prediction remain
unchanged. Only the cells available to the cohort-dependent refinement stage
are modified.

\subsection{Data Processing and Base Classification}
\label{subsec:data_processing}

We use the PBMC3K and Paul15 datasets provided through Scanpy. Classes with
insufficient support are removed, and large classes are capped to reduce class
imbalance and control runtime. We then create stratified training and test
sets.

For nonnegative count data, each cell is library-size normalized and
log-transformed. Highly variable genes are retained, and the selected features
are scaled before dimensionality reduction. Datasets that are already
transformed and contain negative values are not renormalized. For those
datasets, genes are selected according to expression variance. Principal
component analysis is then applied to obtain a lower-dimensional
representation.

We train multinomial logistic regression and a calibrated linear support vector machine (SVM). Both classifiers use class balancing. The linear SVM is calibrated to output class probabilities.

Let $C=|\mathcal{Y}|$ be the number of retained cell types. The frozen base
classifier produces a probability vector
$\mathbf{p}_i=(p_{i1},\ldots,p_{iC})$ for cell $i$, where
$\sum_{c=1}^{C}p_{ic}=1$. Its independent prediction is
$\hat{y}^{\,\mathrm{base}}_i=\arg\max_{c}p_{ic}$.

\subsection{Cohort-Dependent Refinement}
\label{subsec:refinement}

For each test cell, we construct a Euclidean nearest-neighbor graph in the PCA
feature space. Let $\mathcal{N}_k(i)$ denote the indices of the $k$ nearest
test cells to cell $i$, excluding the cell itself.

The confidence of neighbor $j$ is defined as
$q_j=\max_{c}p_{jc}$. For confidence-weighted refinement, the neighborhood
support assigned to class $c$ is

\begin{equation}
h_{ic}=
\frac{\sum_{j\in\mathcal{N}_k(i)} q_j p_{jc}}
{\sum_{j\in\mathcal{N}_k(i)}q_j}.
\label{eq:neighbor_support}
\end{equation}

For uniform voting, all $q_j$ values in
Eq.~\eqref{eq:neighbor_support} are replaced by one. The base probabilities and
neighborhood support are combined as

\begin{equation}
r_{ic}=
\frac{\alpha p_{ic}+\lambda h_{ic}}
{\sum_{\ell=1}^{C}\left(\alpha p_{i\ell}+\lambda h_{i\ell}\right)},
\label{eq:refined_probability}
\end{equation}

where $\alpha$ is the base-prediction weight and $\lambda$ is the context
weight. The refined label is $\hat{y}_i=\arg\max_c r_{ic}$.

After removing $\mathcal{S}$, the base probability vectors of retained cells
are reused without retraining. The neighbor graph and refined probabilities
are recomputed on $\mathcal{D}_{\mathcal{S}}$. This isolates the effect of
query-cohort composition.

\subsection{Target Selection}
\label{subsec:target_selection}

For each dataset, classifier, and random seed, we first identify test cells
whose clean refined prediction is correct. These cells are ranked by their
maximum refined probability. A candidate set is formed from lower-confidence
correctly annotated cells, after which targets are sampled without replacement.

This policy focuses the audit on correctly annotated cells that are more
likely to depend on cohort context. The same target manifests are reused in
follow-up searches and ablations so that methods are compared on matched
targets.

\subsection{Companion-Cell Removal}
\label{subsec:removal_methods}

For a removal fraction $\rho$, the requested removal count is
$b=\max(1,\operatorname{round}(\rho(n_{\mathrm{test}}-1)))$, where
$n_{\mathrm{test}}$ is the test-cohort size.

\subsubsection{Random Removal}

The random baseline samples $b$ non-target cells uniformly without
replacement. Each target-budget combination is evaluated over repeated random
removal sets.

\subsubsection{Same-Class Removal}

Same-class removal forms a pool containing non-target cells whose clean
refined label equals the clean refined label of the target. Up to $b$ cells
are sampled without replacement from this pool. This method removes cells that
support the target's current class, but it does not use distance from the
target.

\subsubsection{Nearest-Cell Removal}

Nearest-cell removal computes the Euclidean distance between the target and
every other test cell in PCA space. The $b$ closest non-target cells are
removed. This method directly changes the target's local neighborhood.

\subsection{Search-Based Removal}
\label{subsec:search_methods}

The search methods optimize removals for one target at a time. Their candidate
pool contains the target's clean first-order neighbors and the neighbors of
those cells. If needed, this set is supplemented with additional cells selected
by distance in PCA space.

For the target's clean refined label, let $c_t$ denote its corresponding class
index. We measure its margin after removing $\mathcal{S}$ as 
$m_t(\mathcal{S})=
r_{t c_t}^{(\mathcal{S})}
-
\max_{c\neq c_t}r_{tc}^{(\mathcal{S})}$, where $r_{tc}^{(\mathcal{S})}$ is the refined probability recomputed on
$\mathcal{D}_{\mathcal{S}}$. A negative margin implies that another class has
overtaken the original class.

Candidate removals are ranked lexicographically. A target flip is preferred
over a non-flip. Among outcomes with the same flip status, a lower value of
$m_t(\mathcal{S})$ is preferred, followed by a lower collateral flip rate.

\subsubsection{Greedy Search}

The candidate ordering is divided into non-overlapping groups. At each step,
every remaining group is temporarily added to the current removal set. The
highest-ranked outcome is retained, and the selected group is removed from the
candidate list. The process stops after a successful flip, after reaching the
search limit, or after exhausting the candidate pool.

\subsubsection{Multi-Start Greedy Search}

The multi-start method repeats the greedy procedure under several reproducible
candidate orderings. A successful trajectory is preferred over an unsuccessful
one. Successful trajectories are then ranked by fewer removed cells, lower
final target margin, and lower collateral damage.

\subsubsection{Beam Search}

Beam search maintains several partial removal sets at each depth. Every retained
state is expanded using a restricted set of unused candidate groups, after
which only the highest-ranked non-successful states are kept. If one or more
expansions flip the target, the method returns the successful path with the
fewest removed cells and the best final score.

\subsection{Attack Success and Collateral Changes}
\label{subsec:evaluation_metrics}

Let $\hat{y}^{(0)}_t$ be the clean refined label of the target and
$\hat{y}^{(\mathcal{S})}_t$ its refined label after removal. The target-flip
indicator is

\begin{equation}
A_t(\mathcal{S})=
\mathbb{I}
\left[
\hat{y}^{(\mathcal{S})}_t
\neq
\hat{y}^{(0)}_t
\right],
\label{eq:attack_success}
\end{equation}

where $\mathbb{I}[\cdot]$ is one when its condition is true and zero otherwise.
The target flip rate is the mean of $A_t(\mathcal{S})$ across evaluated
targets.

To measure unintended effects, we evaluate a subset of retained non-target
cells. Let $\mathcal{E}_{\mathcal{S}}$ denote this evaluation set. The
collateral flip rate is

\begin{equation}
G_t(\mathcal{S})=
\frac{1}{|\mathcal{E}_{\mathcal{S}}|}
\sum_{i\in\mathcal{E}_{\mathcal{S}}}
\mathbb{I}
\left[
\hat{y}^{(\mathcal{S})}_i
\neq
\hat{y}^{(0)}_i
\right].
\label{eq:collateral_rate}
\end{equation}

We also report the removed fraction, target original-class probability, target
margin, clean accuracy, macro-F1, and disagreement between base and refined
predictions.

\subsection{Mechanism Ablations}
\label{subsec:ablations}

We repeat the structured-removal experiments while changing one refinement
setting at a time. The ablations vary the context weight, neighborhood size,
and voting rule. All ablations reuse the targets selected in the main
experiment.

When $\lambda=0$, the model does not use information from neighboring cells. Therefore, removing companion cells cannot change the target prediction.

\subsection{CellTypist Validation}
\label{subsec:celltypist_validation}

We further evaluate the threat model using CellTypist with a pretrained immune
cell model and its built-in majority-voting procedure. PBMC3K is stratified by
its reference labels and reduced to a manageable query cohort. Raw counts are
normalized and log-transformed before annotation.

CellTypist first assigns an independent label to each cell and then performs
over-clustering and cluster-level majority voting. We select cells whose
independent and majority-voted labels disagree, as well as cells whose two
labels agree. These groups are referred to as context-sensitive and initially
stable targets, respectively.

For each target, we remove a small fraction of the remaining cells using
random, nearest-cell, or same-majority-label removal. Nearest cells are
identified in a separate PCA space. After every intervention, CellTypist
reconstructs its over-clustering and majority-voted annotations.

The target expression vector is hashed before and after removal, and execution
stops if the hashes differ. We separately record changes in the independent
CellTypist label and the majority-voted label. This verifies whether a target
flip is caused by cohort-based refinement rather than a change to the target
cell or its independent classifier output.

\section{Results}
\label{sec:results}

This section reports the main removal results, search-based attacks, mechanism
ablations, and CellTypist validation. Experimental settings are given first,
followed by the results for each experiment group.

\subsection{Experimental Setup}
\label{subsec:results_setup}

The target expression vector, reference labels, trained classifier, classifier
probabilities, and model parameters remain unchanged. After each intervention,
only the neighborhood graph and cohort-dependent refinement are recomputed.

We evaluated the controlled pipeline on PBMC3K and Paul15 using multinomial
logistic regression and calibrated linear SVM. Classes with fewer than 30
cells were removed, and each remaining class was limited to 500 cells. We used
a stratified 65--35 training-test split. Nonnegative data were normalized to
$10^{4}$ counts per cell and log-transformed. Up to 1500 genes were retained,
followed by scaling and PCA with 40 components. The main refinement used
$k=25$, with the base and context weights both set to one.

Experiments used seeds 13, 37, and 73. For each dataset, classifier, and seed,
we selected 100 correctly annotated lower-confidence targets, giving 1200
targets in total. Random, nearest-cell, and same-class removal were evaluated
at budgets of 0.5\%, 1\%, 2\%, 5\%, and 10\%. Random removal was repeated ten
times.

Search experiments were performed on Paul15. The candidate pool contained at
most 120 cells and used groups of three. The original greedy search used at
most 15 steps. Search V2 used eight multi-start trajectories, at most 16
steps, and a maximum removal budget of 5\%. Beam search used width four and
expanded each state with up to ten candidate groups.

The ablations used context weights of 0, 0.5, 1, and 2, neighborhood sizes of
10, 25, and 50, and both confidence-weighted and uniform voting. CellTypist
validation used the \texttt{Immune\_All\_Low.pkl} model on 1200 PBMC3K cells
per seed. We selected ten context-sensitive and ten initially stable targets
per seed, giving 60 targets. Random, nearest-cell, and same-majority-label
removal were evaluated at 1\% and 2\% budgets.

The reported flip rates are measured on correctly annotated,
lower-confidence targets selected for robustness analysis. Therefore, they should be interpreted as vulnerability rates within this targeted audit
set, rather than as estimates over all cells in the dataset. The same target
manifests were reused across removal methods, searches, and ablations to
support matched comparisons.

\subsection{Main Companion-Cell Removal Results}
\label{subsec:main_results}

\begin{figure}[t]
    \centering
    \includegraphics[width=\columnwidth]
    {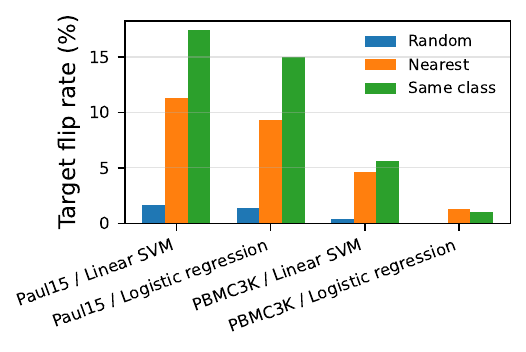}
    \caption{Target flip rates under random, nearest-cell, and same-class
    companion-cell removal at the 5\% budget. Structured removal is strongest
    on Paul15, while PBMC3K is more stable. Results are aggregated across three
    seeds.}
    \label{fig:structured_vs_random}
\end{figure}

Figure~\ref{fig:structured_vs_random} compares the removal strategies at the
5\% budget. Random removal produced low target flip rates in all settings,
while structured removal was substantially stronger on Paul15.

For Paul15 with linear SVM, random removal changed 1.63\% of targets, compared
with 11.33\% for nearest-cell removal and 17.39\% for same-class removal. For
logistic regression, the corresponding rates were 1.37\%, 9.33\%, and
15.00\%.

PBMC3K was more stable. With linear SVM, the flip rates were 0.37\% for random,
4.67\% for nearest-cell, and 5.67\% for same-class removal. With logistic
regression, all rates were at or below 1.33\%.

Paired tests confirmed the Paul15 differences. For linear SVM, nearest-cell
and same-class removal exceeded random removal by 9.70 and 15.82 percentage
points, respectively. For logistic regression, the corresponding differences
were 7.97 and 13.63 points. All four comparisons were statistically
significant. Mean collateral flip rates in the controlled pipeline remained
below approximately 1.1\%.

\subsection{Search-Based Removal}
\label{subsec:search_results}

The original greedy search changed 18.0\% of Paul15 linear-SVM targets and
13.7\% of logistic-regression targets. Increasing the candidate pool and
maximum budget reduced performance, showing that a larger search space does
not always improve a myopic greedy method.

Table~\ref{tab:search_v2} summarizes the Search V2 results. Multi-start greedy
changed 24.33\% of linear-SVM targets and 19.67\% of logistic-regression
targets, while beam search reached 19.33\% and 15.67\%, respectively.

\begin{table}[t]
\centering
\caption{Search V2 results on Paul15. Removal and collateral rates are
calculated over successful attacks.}
\label{tab:search_v2}
\resizebox{\columnwidth}{!}{
\begin{tabular}{llrrr}
\hline
Classifier & Method & Success & Median removal & Collateral \\
\hline
Linear SVM & Beam & 19.33\% & 0.64\% & 0.32\% \\
Linear SVM & Multi-start & \textbf{24.33\%} & 0.64\% & 0.39\% \\
Logistic regression & Beam & 15.67\% & 1.27\% & 0.38\% \\
Logistic regression & Multi-start & \textbf{19.67\%} & 1.27\% & 0.39\% \\
\hline
\end{tabular}}
\end{table}

Successful multi-start attacks required a median removal of 0.64\% for linear
SVM and 1.27\% for logistic regression. As shown in
Table~\ref{tab:search_v2}, mean collateral flip rates remained below 0.4\%,
showing that the selected target could often be changed without broad
disruption.

\subsection{Mechanism Ablations}
\label{subsec:ablation_results}

\begin{figure}[t]
    \centering
    \includegraphics[width=\columnwidth]
    {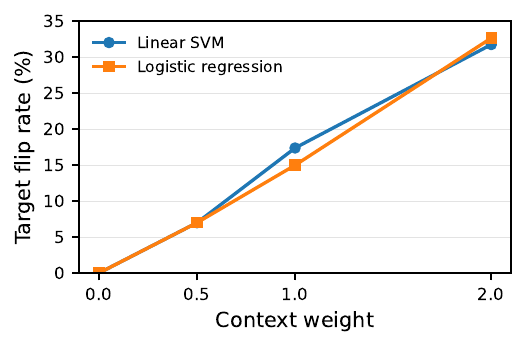}
    \caption{Effect of context weight on target flip rates for same-class
    removal on Paul15. No flips occur when neighborhood refinement is
    disabled, while vulnerability increases as more weight is assigned to
    neighboring cells.}
    \label{fig:context_weight}
\end{figure}

Figure~\ref{fig:context_weight} shows that no target flips occurred when the
context weight was zero. Vulnerability increased with the contribution of
neighboring cells, reaching 31.77\% for linear SVM and 32.67\% for logistic
regression at the largest tested weight. Nearest-cell removal was stronger
with smaller neighborhoods, while same-class removal remained effective
across all tested neighborhood sizes. Uniform voting also remained vulnerable,
showing that the effect does not depend only on confidence weighting.

\subsection{CellTypist Validation}
\label{subsec:celltypist_results}

Independent CellTypist predictions remained unchanged in all 840 evaluations,
so all target flips were confined to majority voting.

Table~\ref{tab:celltypist} summarizes the CellTypist results. Context-sensitive
targets were substantially more vulnerable than initially stable targets. At
the 1\% budget, random removal changed 26.67\% of context-sensitive targets,
while nearest-cell and same-majority-label removal each changed 46.67\%. At the
2\% budget, the corresponding rates were 27.33\%, 40.0\%, and 50.0\%.

Initially stable targets had much lower flip rates, ranging from 3.33\% to
10.0\%. At the 2\% same-majority-label condition, the context-sensitive group
reached 50.0\%, compared with 3.33\% for the initially stable group.

\begin{table}[t]
\centering
\caption{CellTypist majority-voted target flip rates.}
\label{tab:celltypist}
\resizebox{\columnwidth}{!}{
\begin{tabular}{llrrr}
\hline
Target group & Budget & Random & Nearest & Same label \\
\hline
Context-sensitive & 1\% & 26.67\% & 46.67\% & 46.67\% \\
Context-sensitive & 2\% & 27.33\% & 40.00\% & 50.00\% \\
Initially stable & 1\% & 4.00\% & 3.33\% & 6.67\% \\
Initially stable & 2\% & 5.33\% & 10.00\% & 3.33\% \\
\hline
\end{tabular}}
\end{table}

CellTypist produced mean collateral majority-label changes of approximately
10\%. Unlike the controlled pipeline, it reconstructs the full neighborhood
graph and over-clustering after each intervention. Its results therefore show
real query-cohort instability, but not a fully localized attack.

\section{Discussion}
\label{sec:discussion}

CohortHijack shows that clean annotation accuracy does not fully describe the
reliability of cohort-dependent pipelines. A cell may be classified correctly
while its final label still depends strongly on the companion cells present in the query cohort.

The stronger results on Paul15 suggest that this risk is greater when cell
populations overlap or lie along biological transitions. In such settings,
local support may have more influence than in datasets with clearly separated
cell types. This may also help explain why class frequency and clean confidence did not
consistently predict vulnerability.

The search results show that cohort manipulation is not monotonic. Removing
more cells or considering a larger candidate pool can produce a weaker result
because each intervention changes the neighborhood structure. Exploring
multiple search paths is more useful than relying on one greedy
trajectory or a small number of random subsamples.

The CellTypist experiment also reveals a broader reproducibility issue.
Different filtering decisions can rebuild the graph and over-clustering,
leading to different refined labels even when the retained target cells are
unchanged. Annotation tools should therefore report independent and refined
predictions separately and flag cells for which they disagree.

A practical robustness check is to repeat annotation under several controlled
subsamples and report label stability for each cell. Unstable cells could be
assigned a broader lineage label or sent for manual review. This would provide
information that ordinary confidence scores may miss.

The study is limited to two datasets, two linear classifier families, and one
CellTypist model. Future work should examine other annotation systems and
develop refinement methods that preserve the benefits of cohort context while
reducing sensitivity to small cohort changes.

\section{Conclusion}
\label{sec:conclusion}
We introduced CohortHijack, a robustness audit for companion-cell removal in single-cell annotation. Structured and search-based removals changed selected
target labels more often than random removal, while collateral effects remained
limited in the controlled pipeline. The ablations isolated neighborhood
refinement as the source of the vulnerability, and CellTypist validation
confirmed that the effect also appears in an established annotation tool. These findings motivate stability testing as a standard part of evaluating
single-cell annotation systems.

\bibliographystyle{IEEEtran}
\bibliography{IEEEabrv,references}

\end{document}